\documentclass{svproc}
\usepackage{url}
\usepackage{booktabs}
\usepackage{amsmath}
\usepackage{subcaption}
\usepackage{graphicx}
\usepackage{listings}
\usepackage{pdflscape}
\usepackage{textcomp}
\usepackage{multirow}
\usepackage{threeparttable}

\begin{document}
\mainmatter              
\title{Deep Learning-Based Computational Model for Disease Identification in Cocoa Pods (\textit{Theobroma cacao} L.)}
\titlerunning{Computational model for identification of diseases of cocoa pods.}  %

\author{Darlyn Buenaño Vera\inst{1} \and Byron Oviedo\inst{2} \and Washington Chiriboga Casanova\inst{1} \and Cristian Zambrano-Vega\inst{1}\thanks{Corresponding author:  \email{czambrano@uteq.edu.ec}}}

\authorrunning{Zambrano-Vega et al.}

\tocauthor{Darling Buenaño Vera, Byron Oviedo, Washington Chiriboga and Cristian Zambrano-Vega,}

\institute{
	State Technical University of Quevedo, Los Ríos, Ecuador,\\
	Department of Engineering Science \\
	\email{\{darlyn.buenano2017, wchiriboga, czambrano\}@uteq.edu.ec}
	\and
	State Technical University of Quevedo, Los Ríos, Ecuador,\\
	Department of Graduate Programs\\
	\email{boviedo@uteq.edu.ec}
}

\maketitle              

\begin{abstract}

The early identification of diseases in cocoa pods is an important task to guarantee the production of high-quality cocoa. The use of artificial intelligence techniques such as machine learning, computer vision and deep learning are promising solutions to help identify and classify diseases in cocoa pods. In this paper we introduce the development and evaluation of a deep learning computational model applied to the identification of diseases in cocoa pods, focusing on “monilia” and “black pod” diseases. An exhaustive review of state-of-the-art of computational models was carried out, based on scientific articles related to the identification of plant diseases using computer vision and deep learning techniques. As a result of the search, EfficientDet-Lite4, an efficient and lightweight model for object detection, was selected. A dataset, including images of both healthy and diseased cocoa pods, has been utilized to train the model to detect and pinpoint disease manifestations with considerable accuracy. Significant enhancements in the model training and evaluation demonstrate the capability of recognizing and classifying diseases through image analysis. Furthermore, the functionalities of the model were integrated into an Android native mobile with an user-friendly interface, allowing to younger or inexperienced farmers a fast and accuracy identification of health status of cocoa pods.

\end{abstract}
\section{Introduction}

Agriculture is one of the main factors that determines the economic growth of any country. In Ecuador, the majority of the country's agricultural production is carried out by small farmers, representing more than 64\%. Moreover, a significant portion of the food consumed in the country, equivalent to 60\%, comes from peasant family farming \cite{FAO2017}. Within agriculture, cocoa is one of the agricultural products with the highest export volume, with the cocoa sector being the second most exported after bananas. According to the “Ministerio de Producción, Comercio Exterior, Inversiones y Pesca” of Ecuador \cite{MinisterioProduccion2021}, the main destination countries for Ecuadorian cocoa in 2020 were the United States, Indonesia, Malaysia, and the Netherlands. However, due to various seasonal conditions, crops are infected with various types of diseases, making it necessary to take actions for combat and eradication.

Cocoa diseases such as  “monilia” and “black pod” have caused significant economic losses to farming families and companies. Some of these diseases affect the pods directly, while others impact the plantation as a whole. They are responsible for up to 80\% of losses in cocoa production, which can reach 100\% during peak infection periods. “Black pod” is caused by \textit{Phytophthora palmivora} and is one of the most aggressive cocoa diseases worldwide. It produces zoospores that penetrate the plant tissues, leading to the rotting of both the pod and the plant. “Monilia” or “moniliasis” is caused by the fungus \textit{Moniliophthora roreri}. It is an endemic disease that specifically attacks the pods, producing conidia on the infected pod that manifest as wilting, deformities, hydrosis, irregular maturity, necrosis, and oily spots \cite{SolisEtAl2021}. The first case of black pod in Ecuador was reported in 1916, causing annual production losses \cite{MinisterioProduccion2021}.

In this paper, we introduce a deep learning-based computational model leveraging state-of-the-art neural network architectures to accurately identify and classify the “monilia” and “black pod” diseases in cocoa pods. Our approach utilizes a comprehensive dataset of cocoa pod images, annotated with disease markers, to train a model that can diagnose diseases from simple photographs, providing a valuable tool for farmers and agricultural professionals.

\section{Related Works}

Artificial Vision and Deep learning has emerged as a powerful tool in agricultural disease detection, particularly in cacao. Recent studies have focused on developing computational models to accurately identify diseases affecting cacao plants. This section reviews recent advancements in the application of these technologies for disease detection in cacao pods. 

Basri et al. conducted a study comparing various image extraction models for detecting cocoa disease in fruits using Support Vector Machine classification in \cite{Basri2022}. Their research contributes to understanding the effectiveness of different image processing techniques in cocoa disease detection. The classification results using SVM showed the best performance on feature extraction HSV in all types of Kernel SVM used (Linear, RBF, and Polynomial), with the highest accuracy of 80.95\% on RBF Kernel.

A detection of \textit{Phytophthora palmivora} in cocoa fruit with Deep Learning was presented in \cite{detectionPhytophthora2021}. The study utilized the ResNet18 model to detect \textit{Phytophthora palmivora} in cocoa fruits, with a dataset of 1596 images, the model attained an 83\% accuracy in disease detection and 96\% in distinguishing cocoa images from similar fruits.

Kumi et al. developed Cocoa Companion, a smartphone application using deep learning for cocoa disease detection. This app provides farmers with a user-friendly interface to detect diseases in cocoa plants, leveraging state-of-the-art image processing algorithms \cite{Kumi2022}. The automatic detection and diagnosis of diseases is based on the Convolutional Neural Networks (CNN) for image analysis and classification. In the paper, four (4) CNN models are built and trained. The best performing model is SSD MobileNet V2 with over 80\% confidence detection score.

Aubain et al. assessed the fermentation degree of cocoa beans using machine vision. Multi-class support vector machine (SVM) algorithm is used as classifier to discriminate cocoa beans sample into unfermented, partly fermented and well fermented categories. Experimental results show that 99.17\% of UF beans, 97.50\% of PF beans and 100\% of WF beans were detected successfully. This research demonstrates how image processing can be used to evaluate post-harvest processes in cocoa production, contributing to quality control and process optimization. \cite{Aubain2019}.

Finally, Godmalin et al. present a deep learning-based approach to differentiate between healthy and diseased cacao pods. This research highlights the algorithm's accuracy in identifying specific diseases affecting cacao pods. The model can classify three conditions of a given cacao pod image: healthy, black pod disease attack, and pest attack. Under controlled conditions, the model correctly classifies the cacao pod condition with an accuracy of 94\% \cite{godmalin2022classification}

\section{Materials and Methods}

\subsection{Review of Artificial Vision Model Architectures}
\label{sc:review}
Our architectural selection for the computational model was influenced by an extensive review of existing computational models in image recognition. We evaluated seven architectures each with unique strengths and weaknesses, \textbf{AlexNet}: this architecture's large image input size and transfer learning capability seemed promising. However, its high parameter count and substantial computational requirements, along with a lower Top-1 precision of 63.3\% \cite{AlexNetRef}, made it less suitable for our application, \textbf{GoogLeNet}: notable for its application in image classification and object detection, achieved a Top-1 precision of 74.8\% \cite{GoogLeNetRef}. Its limitations in mobile device compatibility and uncertain transfer learning capabilities led us to consider other options, \textbf{ResNet18}: overcoming the gradient vanishing problem and achieving higher precision than AlexNet and GoogLeNet, ResNet18 seemed promising \cite{ResNetRef}. The required conversions for mobile inference, potentially compromising precision, were a concern, \textbf{MobileNetV3-Small}: designed for IoT and mobile devices, offered high precision and parameter efficiency, making it a strong candidate \cite{MobileNetRef}, \textbf{YOLOv3}: known for its robustness in real-time object detection and high precision, YOLOv3's requirement for conversion to mobile-compatible formats was a limiting factor \cite{YOLORef}, \textbf{EfficientDet-Lite4}: tailored for object detection and mobile devices, EfficientDet-Lite4's higher parameter count was a consideration, but its precision was not clearly identified \cite{EfficientDetRef} and \textbf{EfficientNet-Lite4}: excelling in precision, is optimized for mobile devices but has a smaller image input size and is less robust, leading to trade-offs \cite{EfficientNetRef}.

\subsection{Cocoa Pods Image Dataset}
\begin{figure}[!ht]
	\includegraphics[width=12cm]{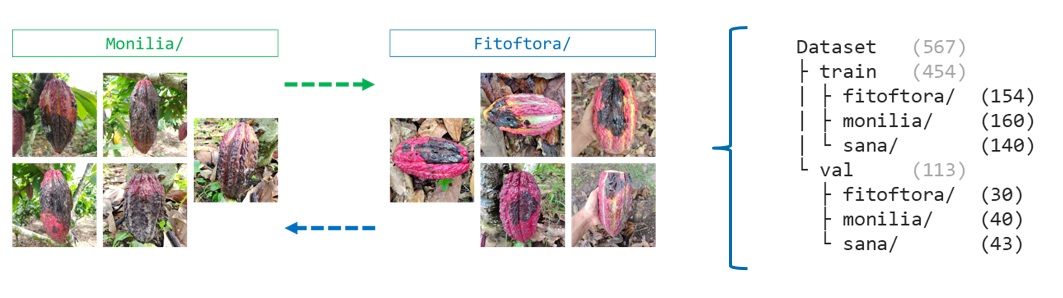}
	\centering
	\caption{Dataset of Cocoa Pods images}
	\label{fig:datasetpreparing}
\end{figure}

The preparation of the data set involved collecting and organizing images of healthy and diseased cocoa pods. These images were either captured by the author or obtained from online platforms, and then divided into training and validation sets (Figure \ref{fig:datasetpreparing}).

\begin{itemize}
\item \textbf{Training Images}
For the training images, photographs of cocoa pods were taken in cocoa plantations in Mocache canton between 13:00 and 15:00 over three days. The camera was maintained at a distance of approximately 40cm from the pod, ensuring that no shadows were cast on the pods due to sunlight. A mobile phone camera was used with the following hardware specifications: 48 MP rear camera with a 4:3 aspect ratio set manually, 4.00 GB RAM, 128GB internal storage, and running on Android 10. The images captured by the author were organized into three directories: one for black pod, another for moniliasis, and the last for healthy pods (Figure \ref{fig:imagedataset}).

\begin{figure}[!ht]
	\centering
	\begin{subfigure}{0.24\textwidth}
		\includegraphics[width=\textwidth]{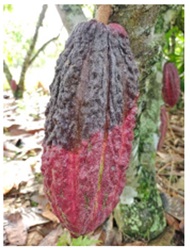}
		\caption{Black pod.}
		\label{fig:blackpod}
	\end{subfigure}
	\hfill
	\begin{subfigure}{0.24\textwidth}
		\includegraphics[width=\textwidth]{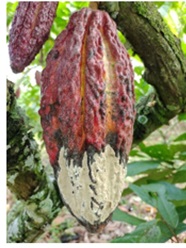}
		\caption{Moniliasis.}
		\label{fig:moniliapod}
	\end{subfigure}
	\hfill
	\begin{subfigure}{0.24\textwidth}
		\includegraphics[width=\textwidth]{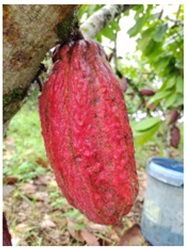}
		\caption{Healthy}
		\label{fig:healthypod}
	\end{subfigure}
	
	\caption{Dataset of Cocoa pods}
	\label{fig:imagedataset}
\end{figure}

\item \textbf{Validation Images}
The validation images were compiled from the online data science community platform Kaggle. A repository titled “Cocoa Diseases (YOLOv4)” was downloaded, containing a dataset of images weighing 2 GiB, classified and labeled into three categories: (a) black pod, (b) moniliasis, and (c) healthy pod.

\end{itemize}

\subsection{Images Preprocessing}
\begin{figure}[!ht]
	\centering
	\begin{subfigure}{0.45\textwidth}
		\centering
		\includegraphics[width=100pt,height=100pt]{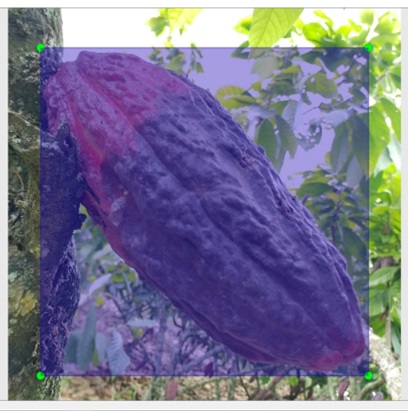}
		\caption{Labeling cocoa pods present in images.}
		\label{fig:labelimg}
	\end{subfigure}
	\hfill
	\begin{subfigure}{0.45\textwidth}
		\centering
		\includegraphics[width=150pt, height=100pt]{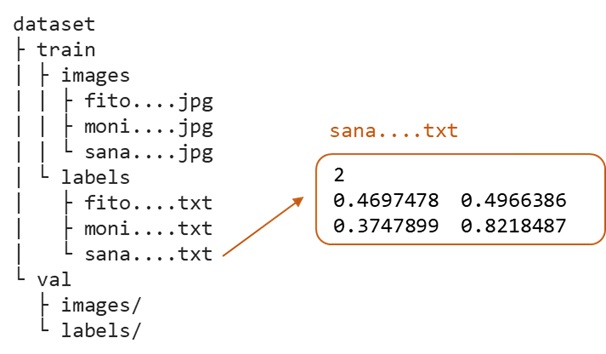}
		\caption{Labels files}
		\label{fig:labelfiles}
	\end{subfigure}
	\caption{Preparation of the images dataset}
	\label{fig:preparationimgs}
\end{figure}

The collected image set underwent data normalization, which involved modifications and transformations to facilitate processing during the training of the machine learning model. Labels were also added to the images to inform the model about the data it was learning. These operations were performed both manually and using automated tools (Figure \ref{fig:preparationimgs}).

\subsection{Data Normalization}
The images in the dataset, originally with a 4:3 aspect ratio, were manually cropped to a square size (1:1 aspect ratio). A Python script was written to resize the images to 640 x 640 pixels, thus reducing the image weight for faster computation. Furthermore, the images retained the three RGB color channels, suitable for the characteristics of cocoa, where the color of the image plays an important role in image recognition.

\subsection{Creation and configuration of the Model}
The computational model was developed in Python using the Google Colab web tool for sophisticated server-based training. We used the TensorFlow Lite's model training tool, Model Maker (TF-Lite Model Maker). The training and testing datasets were located in a Google Drive directory, and the absolute path for each subset was established. This required authorization for Google Colab to access Google Drive. The training configuration primarily involved adjusting two hyperparameters: \textit{epochs} and \textit{batch size}. The epochs refer to the number of times the complete training dataset is used to adjust the model's weights. The batch size indicates the number of training examples used in each model update. We set the model to run for 100 epochs, with a batch size of 17, meaning the model would process 17 image groups per epoch, resulting in 27 steps per epoch. 

Using TF-Lite Model Maker, we created an object detector using the `create` method, which accepts training data, model name, epochs, and batch size as parameters. The model used was `efficientdet\_lite4` from the EfficientDet architecture family for object detection and recognition in images. Two additional parameters were added: one indicating that the model should not be fully trained due to limited server resources, and the second specifying training to commence after the `create` method call. As shown in Listing~\ref{lst:objectdetector}, the object detector model is created using the TensorFlow Lite Model Maker API.

\lstset{ language=Python,	basicstyle=\footnotesize }
\begin{lstlisting}[language=Python, 
	caption={Creation of the model using TFL Model Maker library}, label=lst:objectdetector]
model = object_detector.create(
	train_data,
	model_spec = model_spec.get('efficientdet_lite4'),
	epochs = epochs,
	batch_size = batch_size,
	train_whole_model = False,
	do_train = True,
)

\end{lstlisting}

For exporting the trained model to the TensorFlow Lite format, the `export` method from the TF-Lite Model Maker library was used, allowing the specification of export path and format via parameters. This process took between 40 to 60 minutes.

\subsection{Evaluation Metrics}
The evaluation metrics used in training are described by the following: 

\begin{itemize}
\item \textbf{Classification Accuracy}: Classification Accuracy in object detection models refers to the accuracy of correctly classifying the objects within the bounding boxes, it measures the proportion of correct predictions (both true positives and true negatives) out of all predictions made (Equation \ref{eq:classificationacc}).
\begin{equation}
		Accuracy = \frac{\text{Number of Correct Predictions}}{\text{Total Number of Predictions}}
		\label{eq:classificationacc}
\end{equation}

In terms of true positives (TP), true negatives (TN), false positives (FP), and false negatives (FN), can be defined as follow:

\begin{equation}
	\text{Accuracy} = \frac{TP + TN}{TP + TN + FP + FN}
	\label{eq:accuracy}
\end{equation}

\item \textbf{Precision and Recall}: Precision and Recall are critical metrics in assessing the performance of an object detection model. Are used to measure the performance of a classifier in binary and multiclass classification problems. Precision measures the accuracy of positive predictions, while recall measures the completeness of positive predictions. Both metrics are defined in Equations \ref{eq:precision} and \ref{eq:recall}:

\begin{equation}
	Precision = \frac{TP}{TP + FP}
	\label{eq:precision}
\end{equation}
\begin{equation}
	Recall = \frac{TP}{TP + FN}
	\label{eq:recall}
\end{equation}

\item \textbf{Box (Bounding Box Accuracy)}: Box Accuracy measures the precision of the predicted bounding boxes against the actual (ground truth) boxes. It is commonly quantified using Intersection over Union (IoU), that is calculated between the Area of Overlap between Predicted Box and Ground Truth Box and the Area of Union between Predicted Box and Ground Truth Box (Equation \ref{eq:iou}).
\begin{equation}
	IoU = \frac{\text{Area of Overlap}}{\text{Area of Union}}
	\label{eq:iou}
\end{equation}

\item \textbf{Objectness}: Assess whether a bounding box contains an object. This metric is a binary classification, providing a confidence score of object presence within a bounding box.

\item \textbf{Mean Average Precision (mAP) at IOU thresholds}: mAP at different Intersection Over Union (IOU) thresholds is a common metric in object detection. It is calculated by taking the mean AP over all classes and/or overall IoU thresholds, depending on different detection challenges that exist. It is described in the Equation \ref{eq:map05}: 

\begin{equation}
	\text{mAP} = \frac{1}{N} \sum_{t=1}^{N} \text{AP}_t
	\label{eq:map05}
\end{equation}

Where 
\begin{itemize}
	\item $mAP$ is the mean average precision
	\item $N$ is the number of IoU thresholds
	\item $AP_t$ is the average precision at the $t$-th IoU threshold
\end{itemize}

The Average Precision ($AP$) for each threshold is described in Equation \ref{eq:AP}:

\begin{equation}
	\text{AP}_t = \int_{0}^{1} p(r) \, dr
	\label{eq:AP}
\end{equation}

Where $p(r)$ is the precision at recall $r$.

This integral is typically approximated by averaging precisions at a finite number of equally spaced recall levels.

\item \textbf{mAP@0.5, mAP@0.5:0.95}: These metrics refer to the mean Average Precision (mAP) at different Intersection Over Union (IoU) thresholds:

\begin{itemize}
	\item \textbf{mAP@0.5}: This is the mean Average Precision calculated at an IoU threshold of 0.5. It means that a predicted bounding box is considered a true positive if it has an IoU of 0.5 or more with a ground truth bounding box.
	
	\item \textbf{mAP@0.5:0.95}: This metric averages the mAP calculated at different IoU thresholds, from 0.5 to 0.95, in steps of 0.05. Essentially, it's the average mAP over the range of IoU thresholds between 0.5 and 0.95. The Equation \ref{eq:mAP95} described the formula:
	
	\begin{equation}
		mAP@0.5:0.95 = \frac{1}{10} \sum_{t=0.5}^{0.95} \text{AP@IoU=t}
		\label{eq:mAP95}
	\end{equation}
	
\end{itemize}

\end{itemize}

\subsection{Mobile Application Development}
\begin{figure}[ht]
	\includegraphics[width=7cm]{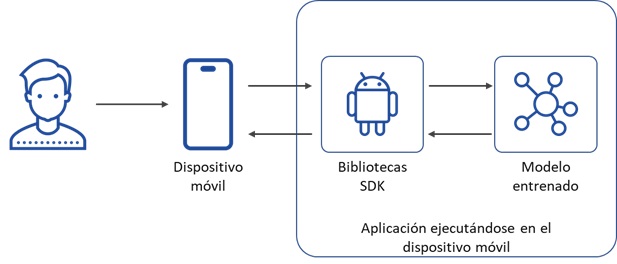}
	\centering
	\caption{Mobile application architecture}
	\label{fig:apparchitecture}
\end{figure}

For the development of the mobile application, software development tools and a methodology were employed to manage the process in several phases. The iterative incremental waterfall model proposed by Winston W. Royce in 1970 was used as the software development methodology \cite{Royce1970}. This model is an adaptation of the traditional waterfall methodology that allows for the review of completed phases, verification of their outcomes, and application of corrections. Brochures on cocoa diseases were examined to define the data to be displayed in the diagnosis of the health state of the cocoa pod. The integration of the trained artificial intelligence model and the interpretation of the data resulting from an input image to the model were scrutinized (Figure \ref{fig:apparchitecture}).

Android Studio was chosen as the Integrated Development Environment (IDE), Java for backend language, XML for designing the user interfaces, and JSON for data structuring. Libraries were imported for using the device's camera and TensorFlow Lite model. Java classes were coded to perform inferences from the trained model, obtain predictions, access the camera and gallery, and save the processed image in the mobile device. The source code was also ensured to meet quality standards and coding best practices.

\section{Results}

\subsection{Selection of the Model Architecture}

Table \ref{tab:neural_network_comparison} contains the computational model architectures identified in the section \ref{sc:review}. The characteristics described were laid out for comparison among the architectures to choose the most suitable one for our research.
\begin{itemize}
	\item \textbf{AlexNet}:
	AlexNet, with its image input size of 227 square pixels, is superior to GoogLeNet, ResNet18, and MobileNetV3-Small in terms of input dimension. It also allows retraining through transfer learning technology. However, it has the highest number of parameters, leading to overfitting with a Top-1 precision of 63.3\%. Its main limitations are its computational expense and gradient vanishing issues \cite{AlexNetRef}.
	
	\item \textbf{GoogLeNet}:
	GoogLeNet is used in both image classification and object detection, achieving a Top-1 precision of 74.8\% and Top-5 precision of 92.2\%. It has limitations in mobile device compatibility and its learning type for retraining is not identified, casting doubt on its adaptability for new tasks \cite{GoogLeNetRef}.
	
	\item \textbf{ResNet18}:
	ResNet18, with its deep layer structure, overcomes the gradient vanishing problem and has fewer parameters (11.7 million) compared to AlexNet and GoogLeNet. It achieves higher precision than these two architectures but requires conversions for mobile inference \cite{ResNetRef}.
	
	\item \textbf{MobileNetV3-Small}:
	MobileNetV3-Small, utilizing AutoML and NetAdapt techniques, has 4.8 million parameters and a Top-1 precision of 69.7\%. It is ideal for mobile devices due to lower computational processing needs and is designed specifically for IoT and mobile devices \cite{MobileNetRef}.
	
	\item \textbf{YOLOv3}:
	YOLOv3 is widely used in object detection, surpassing other architectures in Top-1 and Top-5 precision, and supports larger image inputs. Despite being robust, it requires conversion to mobile-compatible formats like TensorFlow Lite \cite{YOLORef}.
	
	\item \textbf{EfficientDet-Lite4}:
	EfficientDet-Lite4, a recent architecture from the EfficientDet -Lite family, is focused on object detection and designed for mobile devices. However, its precision is not clearly identified, and it has a higher parameter count (15.1 million) compared to YOLOv3 \cite{EfficientDetRef}.
	
	\item \textbf{EfficientNet-Lite4}:
	EfficientNet-Lite4, part of the EfficientNet-Lite family, is optimized for mobile devices, achieving a Top-1 precision of 81.54\% and Top-5 precision of 95.66\%. However, its smaller image input size and non-robust nature make it less performant \cite{EfficientNetRef}.
\end{itemize}

\begin{landscape}
	\begin{table}[t]
		\centering
		\begin{threeparttable}
		
			\begin{tabular}{lccccccc}
				\hline
				\multirow{2}{*}{\centering \textbf{Features}}  &  \multicolumn{7}{c}{\textbf{Architectures}} \\ 
				& AlexNet & GoogLeNet & ResNet18 & MobileNetv3 - Small & YOLOv3 & EfficientDet-L4 & EfficientNet-L4 \\ 
				\hline
				\textbf{Type of tasks} & Img-Clas & Img-Clas and detect & Img-Recog & Img Class & Obj-Detect & Obj-Detect & Img-Clas \\ 
				\textbf{Training data} & ImageNet & ImageNet & ImageNet & ImageNet & COCO & ImageNet & ImageNet \\ 
				\textbf{\#Images Training} & \textgreater{}  1M & \textgreater{} 1M & \textgreater{} 1M & \textgreater{} 1M & \textgreater{} 1M & 25,022 & \textgreater{} 1M  \\ 
				\textbf{\# categories} & \textgreater{} 1K & \textgreater{} 1K & \textgreater{} 1K & 1K & \textgreater{} 1K & \textgreater{} 1K & \textgreater{} 1K \\ 
				\textbf{Input image size (pix)} & 227 x 227 & 224 x 224 & 224 x 224 & 224 x 224 & 640 x 640 & 380 x 380 & 380 x 380 \\ 
				\textbf{Color space} & RGB & RGB & RGB & RGB & RGB & RGB & RGB \\ 
				\textbf{Type of learning} & TF & UnI & UnI & RL & TL & TL & TL \\ 
				\textbf{\# parameters} & 62M & 23M & 11.7M & 4.8M & 36.9M & 15.1M & 13.01M \\ 
				\textbf{Top 1 Accuracy} & 63.30\% & 74.8\% & 72.33\% & 69.7\% & 76.5\% & 57.70\% & 58.80\% \\
				\textbf{Top 5 Accuracy} & 84.60\% & 92.2\% & 91.80\% & Unidentified & 93.3\% & Unidentified & 95.66\% \\ 
				\textbf{Latency} & Not found & Not found & Not found & 15.8 ms & 460 ms & 60 ms & 50 ms \\
				\textbf{Robust network} & Yes & Yes & Yes & No & Yes & No & No \\ 
				\textbf{Model size (MB)} & Not found & Not found & Not found & Not found & 132 MB & 19.9 MB & 15.2 MB \\ \hline
			\end{tabular}
			\begin{tablenotes}
				\item[*] Img-Clas: Image Classification, Obj-Detect: Object Detection, Img-Recog: Image Recognition
				\item[**] UnI: Unidentified, TF: Transfer learning, RL: Reinforcement learning.
			\end{tablenotes}
			\caption{Comparison of different neural network architectures}
			\label{tab:neural_network_comparison}
		\end{threeparttable}
	\end{table}
\end{landscape}

\begin{figure}[!ht]
	\includegraphics[width=12cm]{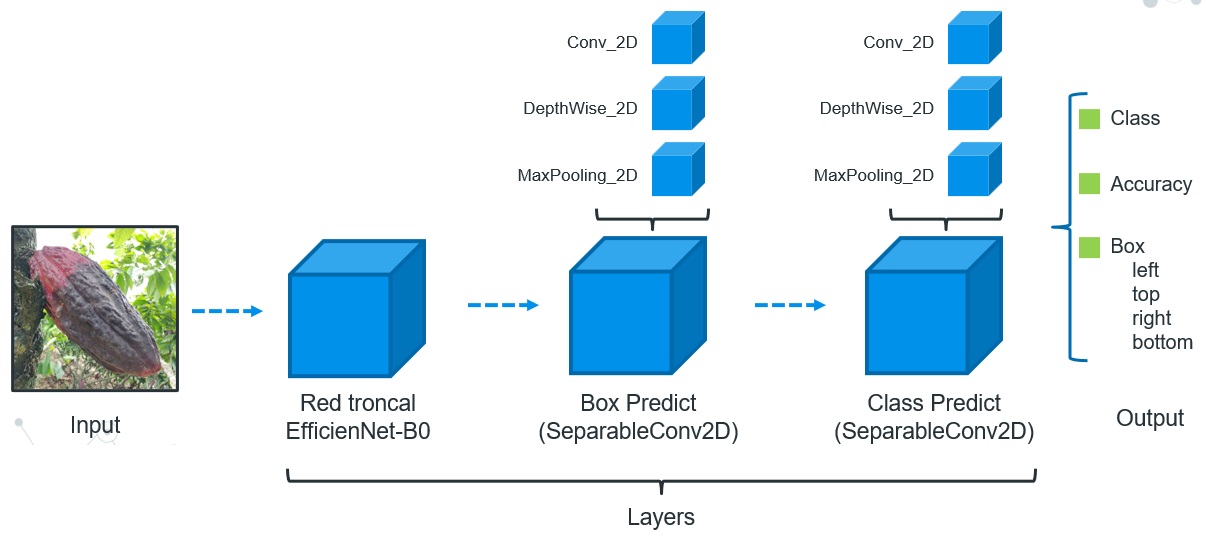}
	\centering
	\caption{Selection of Model Architecture: EfficientDet-Lite4}
	\label{fig:modelarchselected}
\end{figure}

Based on the information in Table \ref{tab:neural_network_comparison}, the most suitable computational model architecture for this project can be selected. The chosen architecture for image recognition tasks is EfficientDet-Lite4 (Figure \ref{fig:modelarchselected}). It has shown acceptable accuracy in category classification and object identification, supporting an input image size of 380x380 pixels, though it can be trained with larger pixel images. Additionally, it is an object detection architecture, meaning it provides information about the object's location within the image. It is also a lightweight and fast network, a useful feature for object recognition in images on limited mobile devices. This choice will enable the development of an effective and efficient system for detecting diseases in cocoa crops.

\subsection{Performance of the Model}
\subsubsection {Results of Model Training and Evaluation}
\begin{figure}[!ht]
	\includegraphics[width=12cm]{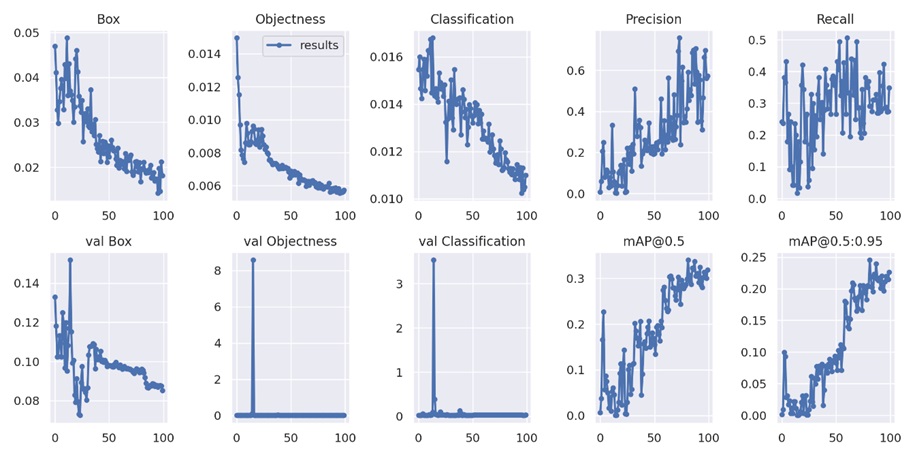}
	\centering
	\caption{Results of the model training}
	\label{fig:resultsmodeltraining}
\end{figure}

The results of the model training are illustrated in Figure \ref{fig:resultsmodeltraining} and indicates the following outcomes based on the presented metrics:

\begin{itemize}
	\item \textbf{Box Loss:} The validation box loss began around 0.05 and showed a decreasing trend, indicating an improvement in the localization ability of the model, settling around 0.02 by the end of 100 epochs.
	\item \textbf{Objectness:} The objectness loss started near 0.012 and significantly decreased, suggesting the model's increased accuracy in predicting the presence of objects, ending just below 0.002.
	\item \textbf{Classification Loss:} The validation classification loss opened at approximately 0.016 and dropped, showing an enhancement in the model's classification capability, concluding around the value of 0.004.
	\item \textbf{Precision:} The precision metric fluctuated considerably with an upward trend, starting from around 0.2 and reaching up to approximately 0.6, indicating the proportion of positive identifications that were correct.
	\item \textbf{Recall:} Recall also varied greatly throughout the training, with values ranging roughly from 0.1 to 0.45, reflecting the model's ability to identify all the relevant instances correctly.
	\item \textbf{mAP@0.5:} Mean Average Precision at an Intersection over Union (IoU) threshold of 0.5 improved steadily from about 0.1 to just over 0.3, indicating better accuracy in the model's object detection over time.
	\item \textbf{mAP@0.5:0.95:} This metric, which is more stringent, shows a consistent increase from near 0 to about 0.23, suggesting the model's detection precision improved across a range of IoU thresholds.
\end{itemize}

The achieved results suggest a progressive enhancement in the model's ability to accurately classify and localize disease conditions in cocoa pods, demonstrating the potential effectiveness of our computational model developed.

\subsubsection {Results of Model Evaluation}
\begin{figure}[!ht]
	\includegraphics[width=10cm]{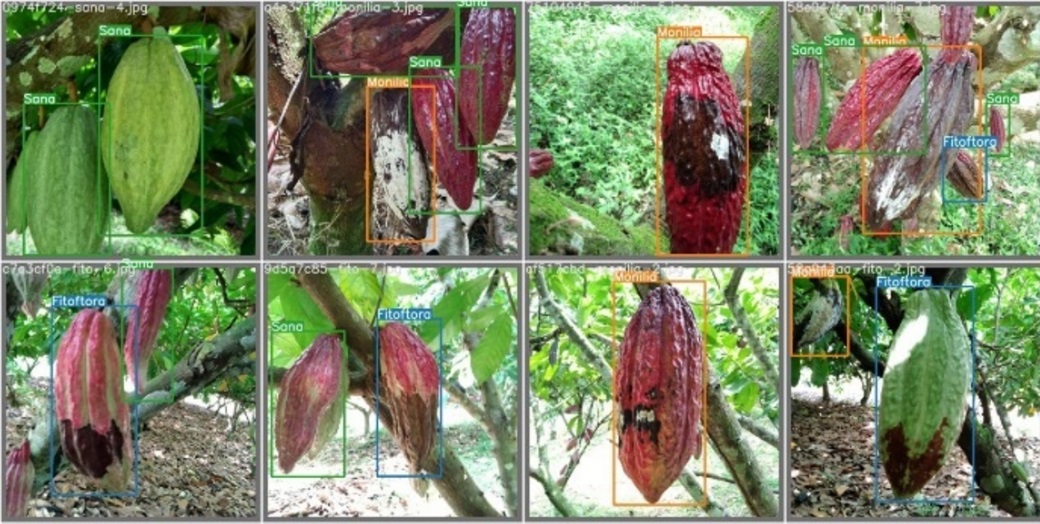}
	\centering
	\caption{Performance of the Model classifying cocoa pods}
	\label{fig:modeleval}
\end{figure}

The model achieved an average precision of 35\%, and a precision of 42.3\% for fitoftora, 27.3\% for monilia, and 34.4\% for healthy ones. Figure \ref{fig:modeleval} illustrates the performance of the model detecting cocoa pods and classifying the healthy and diseased.

\subsection{Cacao DL: A Mobile application for identification of diseases of cocoa pods}

\begin{figure}
	\centering
	\begin{subfigure}{0.30\textwidth}
		\includegraphics[width=\textwidth]{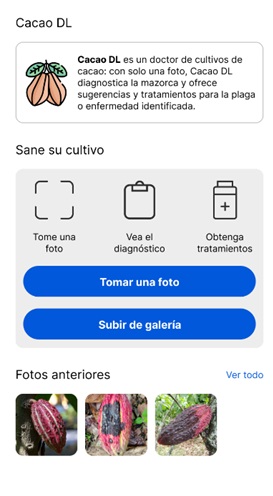}
		\caption{Starting of the app}
		\label{fig:app1}
	\end{subfigure}
	\hfill
	\begin{subfigure}{0.30\textwidth}
		\includegraphics[width=\textwidth]{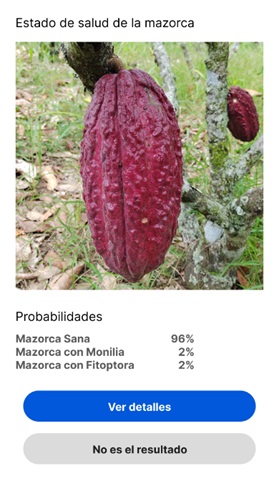}
		\caption{Identification of the disease}
		\label{fig:app2}
	\end{subfigure}
	\hfill
	\begin{subfigure}{0.30\textwidth}
		\includegraphics[width=\textwidth]{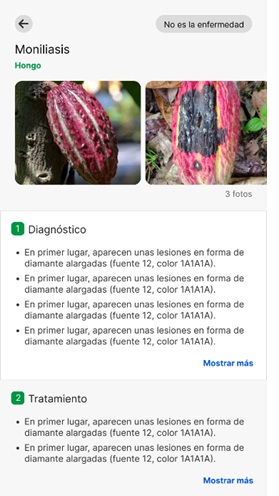}
		\caption{Diagnosis and treatment information}
		\label{fig:app3}
	\end{subfigure}
	
	\caption{Mobile App for identification of diseases of cocoa pods}
	\label{fig:app}
\end{figure}

The Figures \ref{fig:app1}, \ref{fig:app2} and \ref{fig:app3} illustrates the performance of the mobile application developed. The starting screen offers a user-friendly design for diagnosing cocoa pod diseases.

With intuitive icons for taking photos, viewing diagnoses, and obtaining treatment suggestions, the app simplifies disease management for cocoa farmers, integrating the deep learning model for accurate identification and actionable insights. The second image displays the health status of a cocoa pod. For example, below the image, the app provides a probabilistic diagnosis indicating a 96\% likelihood that the cocoa pod is healthy, a 2\% of Monilia infection, and a 2\% of Phytophthora infection. There are two action buttons: "See details," which likely offers a more in-depth analysis, and "Not the result," presumably allowing users to reject the diagnosis if it seems inaccurate. The app's interface presents a diagnosis and treatment section, each with bullet points describing the symptoms and suggested actions. Accompanying the text are images of infected cocoa pods showing characteristic signs of the disease. Users can interact with the "Show more" prompts to expand the information and a button "Not the disease" if the diagnosis does not match the observed symptoms, allowing for user feedback on the accuracy of the app's analysis.

\section{Conclusions and Future Work}

\subsection{Conclusions}
A study of the state of the art was conducted to identify some computational model architectures used in image recognition tasks. Consequently, seven architectures commonly used for detection and classification of images were identified.

EfficientDet-Lite4 was selected as the most suitable computational model architecture compared to others, as it is a fast and lightweight object detector. It also performs satisfactorily in feature extraction from an image and prediction of bounding boxes. Therefore, it was chosen because it is ideal for the model to recognize a cocoa pod and also draw a bounding box around the identified pod.

The dataset taken by the authors required treatment; images were cropped and normalized to create a homogeneous set of images. This process took several days of effort as about 500 images were prepared. Ultimately, the prepared dataset was reviewed and corrected by the “Asociación de Productores Agrícolas Divino Niño” certifying that the images are correctly classified according to health status.

The computational model under the EfficientDet-Lite4 framework was trained for the identification and classification of the health status of cocoa pods using the prepared dataset. The resources of the Colab execution environment were utilized to accelerate the training. Additionally, the loss percentage for classifying health status and delimiting the pods was satisfactory, achieving an accuracy of over 34\%.

A mobile application was successfully developed, enabling users to take photographs of cocoa pods and use the trained model to determine the health status of the pod.

\subsection{Future Work}
Further investigation into more efficient computational model architectures for specific image recognition tasks is recommended. Currently, image recognition is a critical task in many real-life applications, from medical image classification to object detection in autonomous vehicles.

Preparing a larger image dataset is advised to provide the model with a better understanding of the variability present in cocoa pods. The larger the image dataset, the greater the model's ability to recognize important patterns and features in the images, which can in turn improve the accuracy of the predictions made.

Evaluating the scalability and performance of the developed application on various devices of different capabilities, including those that are mid to low-range, is suggested. This will provide insight into the software's behavior under real-world usage conditions and identify potential performance issues.

\bibliographystyle{splncs03_unsrt} 
\bibliography{refs}

\begin{thebibliography}{10}
\providecommand{\url}[1]{\texttt{#1}}
\providecommand{\urlprefix}{URL }

\bibitem{FAO2017}
{Food and Agriculture Organization}: {Ecuador en una mirada}.
  \url{https://www.fao.org/ecuador/fao-en-ecuador/ecuador-en-una-mirada/es/}
  (2017), [Online; accessed 2-July-2022]

\bibitem{MinisterioProduccion2021}
{Ministerio de Producción, Comercio Exterior, Inversiones y Pesca}: {Inició
  Aromas del Ecuador – Edición Cacao, vitrina internacional con compradores
  de tres continentes}.
  \url{https://www.produccion.gob.ec/se-inicio-aromas-del-ecuador-edicion-cacao-vitrina-internacional-con-compradores-de-tres-continentes/}
  (2021), [Online; accessed 30-June-2022]

\bibitem{SolisEtAl2021}
Hidalgo, K.S., Villafuerte, S.P., Coello, D.V., Altuna, J.M.Y., López, L.D.O.:
  {Las enfermedades del cacao y las buenas prácticas agronómicas para su
  manejo}. \url{www.iniap.gob.ec} (2021), [Online; available]

\bibitem{Basri2022}
Basri, et~al.: Comparison of image extraction model for cocoa disease fruits
  attack in support vector machine classification (2022),
  \url{https://doi.org/10.1109/IEIT56384.2022.9967910}

\bibitem{detectionPhytophthora2021}
Montesino, R.Y., Rosales-Huamani, J.A., Castillo-Sequera, J.L.: Detection of
  phytophthora palmivora in cocoa fruit with deep learning. In: Iberian
  Conference on Information Systems and Technologies (2021)

\bibitem{Kumi2022}
Kumi, S., et~al.: Cocoa companion: Deep learning-based smartphone application
  for cocoa disease detection. Procedia Computer Science  (2022),
  \url{https://doi.org/10.1016/j.procs.2022.07.013}

\bibitem{Aubain2019}
Aubain, Y., et~al.: Machine vision-based cocoa beans fermentation degree
  assessment (2019), \url{https://doi.org/10.1007/978-3-030-53187-4_17}

\bibitem{godmalin2022classification}
Godmalin, R.A.G., Aliac, C.J.G., Feliscuzo, L.S.: Classification of cacao pod
  if healthy or attack by pest or black pod disease using deep learning
  algorithm. In: 2022 IEEE International Conference on Artificial Intelligence
  in Engineering and Technology (IICAIET). pp. 1--6. IEEE (2022)

\bibitem{AlexNetRef}
Krizhevsky, A., Sutskever, I., Hinton, G.E.: Imagenet classification with deep
  convolutional neural networks. In: Advances in neural information processing
  systems. pp. 1097--1105 (2012), \url{https://doi.org/10.1145/3065386}

\bibitem{GoogLeNetRef}
Szegedy, C., Liu, W., Jia, Y., Sermanet, P., Reed, S., Anguelov, D., Erhan, D.,
  Vanhoucke, V., Rabinovich, A.: Going deeper with convolutions. In:
  Proceedings of the IEEE conference on computer vision and pattern
  recognition. pp. 1--9 (2015), \url{https://doi.org/10.1109/CVPR.2015.7298594}

\bibitem{ResNetRef}
He, K., Zhang, X., Ren, S., Sun, J.: Deep residual learning for image
  recognition. In: Proceedings of the IEEE conference on computer vision and
  pattern recognition. pp. 770--778 (2016),
  \url{https://doi.org/10.1109/CVPR.2016.90}

\bibitem{MobileNetRef}
Howard, A., Sandler, M., Chu, G., Chen, L.C., Chen, B., Tan, M., Wang, W., Zhu,
  Y., Pang, R., Vasudevan, V., et~al.: Searching for mobilenetv3. arXiv
  preprint arXiv:1905.02244  (2019), \url{https://arxiv.org/abs/1905.02244}

\bibitem{YOLORef}
Redmon, J., Farhadi, A.: Yolov3: An incremental improvement. arXiv preprint
  arXiv:1804.02767  (2018), \url{https://arxiv.org/abs/1804.02767}

\bibitem{EfficientDetRef}
Tan, M., Pang, R., Le, Q.V.: Efficientdet: Scalable and efficient object
  detection. In: Proceedings of the IEEE/CVF conference on computer vision and
  pattern recognition. pp. 10781--10790 (2020),
  \url{https://doi.org/10.1109/CVPR42600.2020.01079}

\bibitem{EfficientNetRef}
Tan, M., Le, Q.V.: Efficientnet: Rethinking model scaling for convolutional
  neural networks. arXiv preprint arXiv:1905.11946  (2019),
  \url{https://arxiv.org/abs/1905.11946}

\bibitem{Royce1970}
Royce, W.W.: Managing the development of large software systems. In:
  Proceedings of IEEE WESCON (1970),
  \url{https://www.cs.umd.edu/class/spring2003/cmsc838p/Process/waterfall.pdf},
  often cited as the first formal description of the waterfall model

\end{thebibliography}

\end{document}